\def\year{2022}\relax
\documentclass[letterpaper]{article} 
\usepackage{aaai22}  
\usepackage{times}  
\usepackage{helvet}  
\usepackage{courier}  
\usepackage[hyphens]{url}  
\usepackage{graphicx} 
\usepackage{natbib}  
\usepackage{caption} 
\DeclareCaptionStyle{ruled}{labelfont=normalfont,labelsep=colon,strut=off} 
\usepackage{algorithm}
\usepackage{algorithmic}

\usepackage{newfloat}
\usepackage{listings}
\floatstyle{ruled}
\newfloat{listing}{tb}{lst}{}
\floatname{listing}{Listing}
\title{Graph Reasoning Networks}
\author{
Markus Zopf$^*$,
Francesco Alesiani$^*$
}
\affiliations{
NEC Laboratories Europe GmbH \\
Kurfuerstenanlage 36, D-69115 Heidelberg \\
}

\usepackage{graphicx}
\usepackage{url}
\usepackage{amsmath}
\usepackage{amsfonts}
\usepackage{amssymb}
\usepackage{amsthm}
\usepackage{xcolor}
\usepackage{mathrsfs}
\usepackage{mathtools}
\usepackage{comment}
\usepackage{subfigure}
\usepackage{gensymb}
\usepackage{amsmath}
\usepackage{amsmath,amsfonts,amssymb}
\usepackage{graphicx}
\usepackage{bm}

\usepackage{threeparttable}
\usepackage{multirow}
\usepackage{booktabs}
\usepackage{tablefootnote}
\usepackage{array}
\usepackage{caption}

\newcommand{\R}{\mathbb{R}}

\newcommand{\grn}{GRN}
\newcommand{\satnettopo}{GRN$_\mathrm{ASC}$}
\newcommand{\satnetgnn}{GRN$_\mathrm{GNN}$}
\newcommand{\satnettopognn}{GRN$_\mathrm{ASC+GNN}$}

\usepackage{mathtools}

\DeclarePairedDelimiter{\diagfences}{(}{)}
\newcommand{\diag}{\operatorname{diag}\diagfences}

\newcommand{\sign}{\operatorname{sign}\diagfences}

\newcommand{\set}{\operatorname{Set}\diagfences}

\newcommand{\encoder}{\mathsf{encoder}}
\newcommand{\reasoner}{\mathsf{reasoner}}

\usepackage{bbm}

\usepackage{mathtools}

\DeclareMathOperator{\binary}{\{0,1\}}

\usepackage{centernot}

\usepackage{wrapfig}
\usepackage{tikz}
\usetikzlibrary{arrows.meta,positioning}

\usepackage[american]{babel}

\usepackage{mathtools} 
\usepackage{amssymb}
\usepackage{booktabs} 
\usepackage{tikz} 

\usepackage{tikz}
\usetikzlibrary{shapes}
\usetikzlibrary{plotmarks}

\newcommand{\triangles}[1]{%
\begin{tikzpicture}[#1]%
\draw (0,0) -- (1ex,1ex);%
\draw (0,0) -- (2ex,0ex);%
\draw (2ex,0ex) -- (1ex,1ex);%
\end{tikzpicture}%
}

\newcommand{\squares}[1]{%
\begin{tikzpicture}[#1]%
\draw (0,0) -- (0ex,1ex);%
\draw (0,0) -- (1ex,0ex);%
\draw (0ex,1ex) -- (1ex,1ex);%
\draw (1ex,0ex) -- (1ex,1ex);%
\end{tikzpicture}%
}

\newcommand{\squaresplus}[1]{%
\begin{tikzpicture}[#1]%
\draw (0,0) -- (0ex,1ex);%
\draw (0,0) -- (1ex,0ex);%
\draw (0,0) -- (1ex,1ex);%
\draw (0ex,1ex) -- (1ex,1ex);%
\draw (1ex,0ex) -- (1ex,1ex);%
\end{tikzpicture}%
}

\newcommand{\connected}[1]{%
\begin{tikzpicture}[#1]%
\filldraw (0,0) circle (2pt);%
\draw (0,0) -- (2ex,0ex);%
\filldraw (2ex, 0) circle (2pt);%
\end{tikzpicture}%
}

\makeatletter
\usepackage{comment}
\let\wfs@comment@comment\comment
\let\comment\@undefined
\usepackage{changes}
\let\wfs@changes@comment\comment
\let\comment\@undefined
\newcommand\comment{%
    \ifthenelse{\equal{\@currenvir}{comment}}
    {\wfs@comment@comment}
    {\wfs@changes@comment}%
}
\makeatother

\usepackage{nicefrac}

\usepackage{amstext} 
\usepackage{array}   
\newcolumntype{L}{>{$}l<{$}} 

\newcommand{\com}[1]{\textcolor{red}{(#1)}} 

\begin{document}

\maketitle

\def\thefootnote{*}\footnotetext{\noindent Equal contribution}\def\thefootnote{\arabic{footnote}}
\def\thefootnote{*}\footnotetext{\noindent Presented at the workshop on  graphs and more complex structures for learning and reasoning at AAAI 2022}\def\thefootnote{\arabic{footnote}}

\begin{abstract}
Graph neural networks (GNNs) are the predominant approach for graph-based machine learning. While neural networks have shown great performance at learning useful representations, they are often criticized for their limited high-level reasoning abilities. In this work, we present Graph Reasoning Networks (GRNs), a novel approach to combine the strengths of fixed and learned graph representations and a reasoning module based on a differentiable satisfiability solver. While results on real-world datasets show comparable performance to GNN, experiments on synthetic datasets demonstrate the potential of the newly proposed method.
\end{abstract}

\section{Introduction}
\label{sec:introduction}
Graph-structured data can be found in a wide range of domains such as computer science (e.g. scientific citation graphs and social networks), chemistry (e.g. molecules), and biology (e.g. proteins). Consequently, graph-based machine learning \cite{Scarselli2009, bronstein2021geometric} has received increasing attention from the machine learning community.

Today, the most popular approaches for graph-based machine learning are different variants of the message-passing approach such as Graph Neural Networks (GNNs) \cite{Scarselli2009}, Graph Convolutional Networks (GCNs) \cite{Kipf2018}, and Graph Attention Networks (GATs) \cite{Velickovic2018}. While these approaches have demonstrated good performance in a wide range of application domains, they are often criticized for their limited high-level reasoning abilities and interpretability \citep{cingillioglu2018deeplogic,garcez2020neurosymbolic,marcus2020decade,chen2020can,loukas2019graph,bodnar2021weisfeiler,dwivedi2021graph}. These shortcomings motivate research to combine the abstraction abilities of continuous neural networks and approaches with better high-level reasoning capabilities. A major challenge in doing so is the fact that many reasoning approaches are not differentiable, which renders an efficient joint training of continuous neural networks reasoning approaches impossible.

In this work, we present Graph Reasoning Networks (GRNs), a novel combination of continuous graph-based machine learning and a logical reasoning approach that can be trained with standard gradient-based optimization methods. GRNs consist of a graph encoding module that maps graphs into a $d$-dimensional feature vector in $[0,1]^d$ and a differentiable satisfiability solver \citet{wang2019satnet} that learns logical rules based on the obtained representation. We propose a semi-learnable encoder that consists of a fixed and a learnable component to combine the strength of previously observed usefulness of fixed encodings with the adaptiveness of learned representations.

Our experiments demonstrate the potential of the newly proposed method as it can learn challenging synthetic datasets that standard GNNs cannot learn without a reasoning module. Furthermore, the proposed approach performs on par on real-world datasets with and without using node features.

\section{Related Work}
\paragraph{GNN and Beyond GNN} GCN \citep{kipf2016semi}, GAT \citep{velivckovic2017graph} or in general neural message passing \citep{gilmer2017neural} are among various commonly used methods to process features over graphs.  PPGN (Provably Powerful Graph Network) \citep{maron2019provably} is a class of Neural Networks that improves the expressiveness of GNN. \cite{morris2019weisfeiler} extends message passing to include higher-order node tuples. However, these methods remain highly limited regarding memory and computational complexity.

\paragraph{Learning Rules over Knowledge Graphs}
Learning rules over Knowledge Graphs (KGs) \citep{qu2020rnnlogic} extracts rules from triplets {\it head}-{\it relationship}-{\it tail} (i.e. $(h,r,t)$), to predict the relationship ($r$) not observed or to predict elements ($h$ or $t$) missing in the KG. These methods do not extend to graph classification.

\paragraph{Differentiable SAT} Differentiable SAT methods \citep{wang2019satnet,wang2019low,wang2017mixing} compute gradients on the rules and on the input and output variables to propagate gradients, thus enabling SAT model to be integrated into machine learning systems. Alternative approaches \citep{cingillioglu2018deeplogic,evans2018learning}, restrict the class of rules to predefined templates. These methods do not consider graph classification.

\section{Graph Reasoning Networks}
\label{sec:method}
In the following, we describe our novel approach to combine graph-based machine learning with a reasoning module to mitigate the limited reasoning capabilities of graph neural networks. On a high-level view, the proposed method consists of two submodules: an $\encoder$ and a $\reasoner$. The encoder is a module that takes a graph  $g$ as input and generates a $d$-dimensional intermediate vector representation $r$ of the graph. The intermediate vector representation $r$ contains binary values (i.e. $r_i \in \{0, 1\}$) and/or probabilistic values (i.e. $r_i \in [0, 1]$). In contrast to previous approaches, we propose an encoder that consists of both learnable and not learnable features to empower the reasoner to learn even better rules (hence, a \emph{semi-learnable} encoder). The reasoner is a module that learns logical rules based on the semi-learned $d$-dimensional intermediate representation $r$ and generates a task-specific prediction $o$ for the input graph as an output. Both encoder and reasoner can be trained end-to-end with gradient-based optimization thanks to the differentiable MAX-SAT relaxation proposed in \cite{wang2017mixing}, which is a crucial property of the proposed method. Note that different to most neural networks, the output of the reasoner is not a probability distribution over all possible classes, but a discrete output representing the corresponding class. Hence, for a binary classification problem, we obtain $o \in \{0,1\}$ and a corresponding encoding for multi-class prediction problems.

Based on the two submodules, Graph Reasoning Networks can be defined as a combination of both functions according to
\begin{align*}
    r &= \encoder(g), & y &= \reasoner(r)
\end{align*}
In the following, we discuss both encoder and reasoner in detail.

\begin{comment}

\begin{figure}[h]
\includegraphics[width=0.5\textwidth]{figures/GraphSAT2.png}
\caption{Combining topological and static node features}
\label{fig:GRN_static}
\end{figure}

\end{comment}

\subsection{Semi-learnable encoder}
\label{sec:static_encoder}
To combine the strengths of fixed features and learnable representations, we propose a semi-learnable encoder that maps input graphs into a vector representation, which will be used by the subsequent reasoner to predict a classification output.

\subsubsection{Fixed features.}

The first set of functions consists of fixed, predefined features that encode information about the topology of the graph and the node features (if present). A simple approach to encode the topology as a vector is to flatten the corresponding adjacency matrix $A$ into an adjacency string $S$. To this end, $A \in \binary^{n \times n}$ is converted into
$S \in \binary^{n^2}$ according to $S_{i + n \cdot (j-1)} = A_{i,j}$ for $i,j \in \{1,\dots , n\}$. The size of the adjacency matrix increases quadratically with the number of nodes in the graph, which is also true for the adjacency string $S$.
However, in many datasets such as NCI1 and PROTEINS, the number of nodes in the graphs is rather small and thus allows an application of this approach. Furthermore, in undirected graphs, only a part of the adjacency matrix needs to be encoded since it already contains all information about the graph topology. Moreover, the elements $A_{i,i}$ do not have to be encoded in $S$ if the graphs do not contain self-loops. Hence, the size  of the adjacency string $S$ can be reduced to $l = \frac{(n-1) n}{2}$.

A problem with this approach is that isomorphic graphs with different adjacency matrices $A$ and $A'$ are mapped to different adjacency strings. Two graphs are called isomorphic if there exists an {\it isomorphism} (i.e. a bijection map) between their vertex sets that preserves the graph topology. Graph {\it canonical labelling} \citep{mckay2014practical} is the process where a graph is relabelled such that isomorphic graphs are identical after relabelling. While the complexity of graph canonical labelling is $e^{O(\sqrt{n \log n})}$ \citep{babai1983computational}, there exist fast implementations such as {\tt Trace} or {\tt nauty} \cite{MCKAY201494}.

\begin{comment}
Another popular approach to encode the topology graphs is based on the (1-dimensional) Weisfeiler-Leman graph isomorphism test (1-WL). 1-WL tests if two graphs are isomorphic by iteratively relabeling the nodes in the graph. In each iteration, a histogram of the newly assigned labels is generated and compared with the histogram of the other graph. If the two histograms differ, it is guaranteed that the graphs are not isomorphic. The algorithm stops after fixed number of predefined iterations, when the histograms of the graphs do not change anymore, or when a difference between the graph histograms has been detected. This histogram generation process can be used as a basis to generate a graph representation by simply concatenating the histograms generated in all iterations. Machine learning methods such as SVMs have demonstrated good performance based on the 1-WL representing \com{TODO: cite TUDortmund paper}.
\end{comment}

\begin{comment}
Another interesting choice for a fixed representation function are motif detectors. These are interesting since motifs can be a relevant feature for graph classification tasks. However, the number of possible motifs grows very fast when the size of the motif is increased \com{TODO: how fast?}. A potential solution to this problem is to obtain a set of relevant motifs for specific datasets. For instance, a frequent motif miner can be used to obtain a set of the $n$ most frequent motifs.
\end{comment}

Besides encoding information about the topology, information about the node features can be encoded in a vector representation, for instance by simply concatenating all node features. However, this approach can lead to a large representation, depending on the number of node features.

A problem with the previously described approach is that the rules learned by \citet{wang2019satnet} are not invariant to permutations of the bit string. Hence, each rule is only able to consider a specific sub-structure of a graph or a specific set of node features, which may limit the generalizability of the approach. Furthermore, the representation is fixed and cannot be automatically adapted by a machine learning model based on the concrete problem at hand.


\subsubsection{Learned representation}
\label{sec:learned_encoder}
Encoding the topology with fixed representations such as a topology string or a Weisfeiler-Leman-based representation \cite{Weisfeiler1968} have been shown to be strong features \cite{Shervashidze2011}. However, there are also problems as discussed above. A potential solution is to learn a fixed-sized permutation invariant encoding of the graph. To this end, permutation invariant graph neural networks (GNN) such as GCN \citep{kipf2016semi} or GAT \citep{velivckovic2017graph} can be used. Since the approach proposed by \citet{wang2019satnet} provides gradients not only for the rules but also for the input, the GNN can be trained jointly with the differentiable satisfiability solver such that it learns to generate a useful intermediate representation.

\begin{figure}[h]
\includegraphics[width=0.45\textwidth]{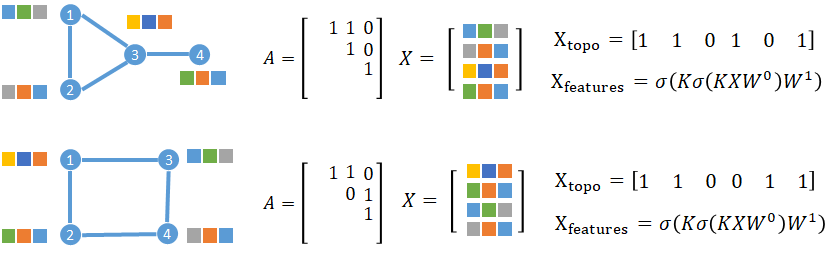}
\caption{Combining topological and learned node features}
\label{fig:GRN_learned}
\end{figure}

\subsubsection{Combining static and learned encoders}
A third option leverages both of the two previously proposed ideas by combining a fixed graph representation with a learned graph representation (see Figure~\ref{fig:GRN_learned}). To this end, the fixed graph representation can simply be concatenated with the learn graph representation. While we do not backpropagate gradients to the fixed graph representation, we can still backpropagate gradients to the GNN to train it.

\subsection{Reasoning module}
The reasoning module takes the previously generated vector representation as input and generates a binary classification as output. The reasoning module is based on a differentiable satisfiability solver proposed by \cite{wang2017mixing} and is explained in detail below.

\subsubsection{Background: Differentiable Satisfiability}

\subsubsection{MAX-SAT}
In Maximal Satisfiability Problems (MAX-SAT), we are interested in finding the assignment of $m$ binary variables $x_i \in \{-1,1\}, i = 1, \dots, m$ for some given rules, or
\begin{align} \label{eq:maxsat0}
    \max_{x \in \{-1,1\}^{m} } & \sum_{j \in [n]} \langle s_{j}, x \rangle
\end{align}
where $s_{ji} \in \{-1,0,+1 \}$ are the rules of the MAX-SAT problem and $ \langle s_{j}, x \rangle  = \lor_{i \in [m]} 1_{ s_{ji} x_i  > 0 }$. MAX-SAT is the relaxation of the Satisfiability (SAT) problem, where all the rules need to be true. The relaxation is useful, when the set of rules may generate an infeasible set of solutions and we are interested in knowing the solution that maximal satisfies the given rules.
\subsubsection{SatNet: Differentiable MAX-SAT relaxation via Semi-definitive programming (SDP)}
Here we consider the relaxation of the MAX-SAT problem as a SDP, \citep{wang2019satnet,wang2019low,wang2017mixing}
, where the MAX-SAT problem defined in Eq.( \ref{eq:maxsat0}) is relaxed by
\begin{align}  \label{eq:maxsat_sdp}
	\min_{V \in \R^{K \times (k+1) } } & \langle S^T S, V^T V \rangle
	\\
	\text{s.t.} & \| v_i \| = 1, \forall i \in \{ \top, 1, \dots, k \}
\end{align}
where for each input variable $x_i$ we have an associated unitary vector $v_i \in \R^K$, with some $K > \sqrt{2k}$ \citep{pataki1998rank}.\footnote{$K$ is the size of the embedded space, while $k$ is the number of variables.} The special variable $v_\top$ represents the true logic value and is used to retrieve the logic value of the solution.  $S =[s_\top,s_1,\dots,s_k]/\diag{1/\sqrt{4|s_j|}} \in \R^{m \times (k+1)}$ is the matrix representing the rules, while $V \in \R^{K \times (k+1)}$ is the matrix of the unitary vectors.
\subsubsection{Mapping from the vector to  the logic variable}
The logic value is recovered by its probability defined as $P\{x_i\} = \frac{1}{ \pi } \arccos {(-v_i ^T v_\top)}$.
The probability measures the radiant angle between the true vector and the variable vector, i.e. $v_i^T v_\top = \cos (\pi x_i)$. We can recover the discrete value by evaluating the sign, i.e. $x_i = \sign {P\{x_i\}}$.
\subsubsection{Mapping from the logic variable to the vector}
To generate the vectors from the logical values, we could use the following mapping $v_i = - \cos (\pi x_i) v_\top + \sin (\pi x_i) P_\top v_i^\text{rand}$,
where $P_i = I_K-v_i v_i^T$ is the projection matrix on the vector $v_i$ and $v_i^\text{rand}$ is a random unit vector.
\subsubsection{Solution to the SDP relaxation}
The solution of Eq.(\ref{eq:maxsat_sdp}) is given by the fix point \citep{wang2019satnet} $v_i  = - \nicefrac{g_i}{\| g_i\|}$,
where $g_i = V S^T s_i - \| s_i \|^2 v_i  = V S^T s_i - v_i  s_i^T s_i$.
\begin{comment}
We can write this in matrix form
\begin{align}
	V = - G \diag { 1 / \| g_i \|} \\G = V [S^TS-\diag{S^TS}]
\end{align}
\end{comment}

\section{Experiments}
\label{sec:experiments}
To evaluate our proposed method, we perform graph classification experiments on synthetic without node features and real-world datasets with and without node features. For the graph classification tasks, we compute the mean prediction accuracy across all graphs in an unseen test set. In all experiments, we report the average result of three runs with three different random seeds to obtain more stable results. To better understand the robustness of the models, we also report the standard deviation (indicated by the $\pm$ symbol). Since we want to evaluate the potential benefits of encoding the topology into a fixed-sized bit string as described above, we filter graphs with a size larger than 15 and 20 nodes. As a consequence, our results are not directly comparable to prior works. Dataset details can be found in Table~\ref{tab:datasets}.

\subsection{Architectures}
Similar to prior works \cite{zhang2019}, we use a graph neural network with 2 convolutional layers with an optional dropout layer after each convolution. A mean pooling layer is used after the convolutional layers to aggregate the obtained node features into a single vector that represents the entire graph. The pooling is followed by an additional layer to map the obtained intermediate representation to the final output. In contrast to our approach, which outputs only a single binary label, the GNN generates two outputs\footnote{As typically done for graph classification, one output for each class: $\{0,1\}$.} for a binary classification task that indicates the probability of each class.

We implement three different versions of the proposed approach based on the previously presented fixed and learned graph representations. The first version, \satnettopo{} (adjacency string canonicalized) uses only the canonicalized adjacency string as graph representation.
Second, we use a version that jointly learns the \grn{} and the GNN, which we denote as \satnetgnn{}. Third, the architecture that uses a combination of both representations is denoted by \satnettopognn{}. To obtain a meaningful comparison with the reference model, we use the same GNN architecture as described above. Instead of using an additional layer to make the class predictions, we use the reasoning module.

\subsection{Hyperparameter optimization}
To perform a hyperparameter optimization, we split the datasets into training, validation, and test splits with sizes of 80\%, 10\%, and 10\% of the dataset, respectively, and report the result of the configuration with the best validation result for each run. For the GNN, we consider a hidden size in $\set{32, 64}$, a learning rate in $\set{0.01, 0.001}$, and test a dropout probability in $\set{0.0, 0.3}$, where a dropout probability of $0.0$ means that no dropout is used. For the Satnet, we consider a learning rate in $\set{0.1, 0.01}$ and a number of rules $m$ and auxiliary variables $aux$ in  $\set{32, 64}$. To limit the search space, we only consider configurations with $m=aux$. Adam optimizer is used to train all models.

\begin{table}
    \centering
	\begin{tabular}{c c c c}
		Dataset & Train & Test  & Num. nodes \\
		\toprule
		NCI1 & 240 & 26 & up to 15 \\
		PROTEINS & 360 & 40 & up to 20 \\
		IMDB-BIN & 400 & 40 & up to 15 \\
	\end{tabular}
	\caption{Dataset statistics}
	\label{tab:datasets}
\end{table}

\begin{table*}
    \centering
	\begin{tabular}{r c L L L}
Model					& Node Features	&	\text{NCI1}				&	\text{PROTEINS}			& 	\text{IMDB-BIN}	\\
\toprule
GNN						& constant		&	{\bf 0.87} \pm 0.02	&	0.63 \pm 0.05 		&	0.54 \pm 0.09	\\
GNN						& node degree	&	0.86 \pm 0.02		&	0.60 \pm 0.07 		&	0.64 \pm 0.05	\\
\satnettopo{}       	&	-			&	{\bf 0.87} \pm 0.11	&	{\bf 0.67} \pm 0.10 &	0.61 \pm 0.07	\\
\satnetgnn{}        	& constant		&	{\bf 0.87} \pm 0.02	&	0.61 \pm 0.05 		&	0.48 \pm 0.00	\\
\satnetgnn{}      	  	& node degree	&	0.83 \pm 0.02		&	0.61 \pm 0.03 		&	{\bf 0.67} \pm 0.05	\\
\satnettopognn{}	 	& constant		&	0.83 \pm 0.09		&	0.63 \pm 0.13 		&	0.63 \pm 0.06	\\
\satnettopognn{} 	    & node degree	&	0.80 \pm 0.07		&	0.62 \pm 0.08 		&	0.62 \pm 0.05	\\
	\end{tabular}
	\caption{Prediction accuracy and standard deviation of three runs for real-world graphs without node features. Column 'Node Feature' indicates which alternative feature has been used as input for the message-passing algorithm. Since \satnettopo{} only uses the topology string, it does not need alternative node features.}
	\label{tab:results_without_node_features}
\end{table*}

\section{Results}
In the following, we report the results on three different types of datasets. We report results on synthetic datasets to illustrate the benefits of the proposed method in challenging problems that cannot be solved by standard GNNs alone. We also report results on real-world datasets with and without node features.

\subsection{Synthetic datasets}
To compare the expressiveness of GRN versus GNN, we generated synthetic graph datasets. We randomly generated graphs with $n$ nodes. We used regular random graphs of fix degree $d$ ($d$-regular) and Erdos-Renyi with edge probability $p$ \citep{bollobas2001random}. As prediction tasks we considered detecting the connectivity of the graph (\connected{}), detecting the presence of motifs: triangles (\triangles{}), squares (\squares{}) and $5$-edges $4$-nodes motif (\squaresplus{}). For $3$-regular graphs we used $th_{\triangles{}}=2, th_{\squares{}}=3$, while we used $th_{\triangles{}}=6, th_{\squares{}}=6,th_{\squaresplus{}}=3$. As expected (see Table~\ref{tab:results_synthetic_datasets}), GNN is not able to detect with accuracy the presence of specific motifs in the graph. The GNN shows a more reasonable performance on the connectivity test, probably exploiting other correlated information. On the other hand, GRN exhibits superior performance, thus confirming that the use of topological information is necessary if the prediction task involves information related to the topology of the graph.

\begin{table}[h]
    \centering
	\begin{tabular}{p{1cm}p{.6cm}p{.6cm}p{.6cm}p{.6cm}p{.6cm}p{.6cm}}
		Dataset & \multicolumn{4}{c}{Erdos-Renyi RG}  & \multicolumn{2}{c}{$3$-Regular RG}\\
        Model & \connected{} & \squares{} & \squaresplus{} & \triangles{} & \squares{} & \triangles{} \\
		\toprule
        GNN		&	0.70	&	0.51	&	0.53	&	0.57	&	0.63	&	0.59 \\
        \satnettopo{}	&	{\bf 0.98}	& {\bf 0.81}	&	{\bf 0.85}	&	{\bf 0.87}	&	{\bf 1.00}	&	{\bf 1.00} \\
	\end{tabular}
	\caption{Results for synthetic graphs with Random Graphs (RG). Prediction tasks: \connected{} for connectivity, \squares{} for square motif counting, \squaresplus{} for $5$ edges motif counting and \triangles{} for triangle counting.}
	\label{tab:results_synthetic_datasets}
\end{table}

\begin{table}
    \centering
	\begin{tabular}{r L L L}
        Model					&	\text{NCI1}	&	\text{NCI109}	&	\text{PROTEINS}	\\
        \toprule
        GNN						&	{\bf 0.88} \pm 0.04	&	{\bf 0.83} \pm 0.06	&	0.60 \pm 0.04 		\\
        \satnetgnn{}        	&	0.87 \pm 0.06	&	0.82 \pm 0.02	&	0.62 \pm 0.06 		\\
        \satnettopognn{} 	    &	0.86 \pm 0.04	&	0.79 \pm 0.06	&	{\bf 0.65} \pm 0.11 		\\
    \end{tabular}
	\caption{Results for real-world graphs with node features.}
	\label{tab:results_with_node_features}
\end{table}

\subsection{Real-world datasets without node features}
Next, we perform experiments on real-world datasets without node features. To this end, we use the NCI1 and the PROTEINS datasets without node features. Furthermore, we use the IMDB-BIN dataset. Since message-passing neural networks such as the GCN rely on node features for message passing, we use two different node feature alternatives. In the first version, we initialize all nodes with the same, constant value. In the second version, we initialize the feature vector of all nodes with their node degree in a one-hot encoding. Using a one-hot representation of the node degree is a strong, hand-crafted feature for GNNs in many datasets. The results of this experiment can be found in Table~\ref{tab:results_without_node_features}.

The results show that the \satnettopo{} and \satnetgnn{} are able to outperform the baseline approaches in the PROTEINS and the IMDB-BIN datasets. Interestingly, \satnettopo{} that does not use the node degree as a feature performs best in PROTEINS, which suggests that the topology is highly informative in this dataset. In NCI1, several methods show a similar performance and \satnettopognn{} does not perform well.

\subsection{Real-world datasets with node features}
In the last experiment, we evaluate the performance of different approaches in the NCI1, NCI109, and PROTEINS datasets with their original node features. The results in Table~\ref{tab:results_with_node_features} show that the baseline GNN performs best in the NCI1 and NCI109 datasets, closely followed by \satnetgnn{}. Additionally using the topology in the \satnettopognn{} seems not to be beneficial in these two datasets. However, \satnettopognn{} performs best in the PROTEINS dataset, which suggests that the model can leverage the information contained in the topology string. This observation confirms the result from Table~\ref{tab:results_without_node_features}, which also showed that the topology seems to be important in the PROTEINS dataset.

\section{Conclusions}
In this work, we present an approach to applying Satnet to graphs. Our experiments on synthetic datasets show that the approach can learn meaningful rules. Moreover, we find that rather simple \satnettopo{} outperforms the GNN baseline. Experiments on real-world datasets with and without node features are less conclusive.

In future work, other fixed representations can be used instead of the adjacency string, which limits the applicability of the approach to small and medium-sized graphs, or as an additional feature of the graph size is not too large. A promising idea is a representation based on the 1-WL graph isomorphism test. Furthermore, we noticed that jointly training the GCN and the SatNet can be difficult in practice because both networks require a different learning rate. Hence, it will be important to develop a robust training schedule to improve the training of the joint model. Inspiration for this can be gained from training procedures that have been developed for GANs that also consist of two modules that depend on each other.



\bibliography{graphsat}

\begin{thebibliography}{29}
\providecommand{\natexlab}[1]{#1}

\bibitem[{Babai, Kantor, and Luks(1983)}]{babai1983computational}
Babai, L.; Kantor, W.~M.; and Luks, E.~M. 1983.
\newblock Computational complexity and the classification of finite simple
  groups.
\newblock In \emph{24th Annual Symposium on Foundations of Computer Science
  (Sfcs 1983)}, 162--171. IEEE.

\bibitem[{Bodnar et~al.(2021)Bodnar, Frasca, Otter, Wang, Li{\`o}, Montufar,
  and Bronstein}]{bodnar2021weisfeiler}
Bodnar, C.; Frasca, F.; Otter, N.; Wang, Y.~G.; Li{\`o}, P.; Montufar, G.~F.;
  and Bronstein, M. 2021.
\newblock Weisfeiler and lehman go cellular: Cw networks.
\newblock \emph{Advances in Neural Information Processing Systems}, 34.

\bibitem[{Bollob{\'a}s and B{\'e}la(2001)}]{bollobas2001random}
Bollob{\'a}s, B.; and B{\'e}la, B. 2001.
\newblock \emph{Random graphs}.
\newblock 73. Cambridge university press.

\bibitem[{Bronstein et~al.(2021)Bronstein, Bruna, Cohen, and
  Veli{\v{c}}kovi{\'c}}]{bronstein2021geometric}
Bronstein, M.~M.; Bruna, J.; Cohen, T.; and Veli{\v{c}}kovi{\'c}, P. 2021.
\newblock Geometric deep learning: Grids, groups, graphs, geodesics, and
  gauges.
\newblock \emph{arXiv preprint arXiv:2104.13478}.

\bibitem[{Chen et~al.(2020)Chen, Chen, Villar, and Bruna}]{chen2020can}
Chen, Z.; Chen, L.; Villar, S.; and Bruna, J. 2020.
\newblock Can graph neural networks count substructures?
\newblock \emph{arXiv preprint arXiv:2002.04025}.

\bibitem[{Cingillioglu and Russo(2018)}]{cingillioglu2018deeplogic}
Cingillioglu, N.; and Russo, A. 2018.
\newblock Deeplogic: Towards end-to-end differentiable logical reasoning.
\newblock \emph{arXiv preprint arXiv:1805.07433}.

\bibitem[{d'Avila Garcez and Lamb(2020)}]{garcez2020neurosymbolic}
d'Avila Garcez, A.; and Lamb, L.~C. 2020.
\newblock Neurosymbolic AI: The 3rd Wave.
\newblock arXiv:2012.05876.

\bibitem[{Dwivedi et~al.(2021)Dwivedi, Luu, Laurent, Bengio, and
  Bresson}]{dwivedi2021graph}
Dwivedi, V.~P.; Luu, A.~T.; Laurent, T.; Bengio, Y.; and Bresson, X. 2021.
\newblock Graph Neural Networks with Learnable Structural and Positional
  Representations.
\newblock \emph{arXiv preprint arXiv:2110.07875}.

\bibitem[{Evans and Grefenstette(2018)}]{evans2018learning}
Evans, R.; and Grefenstette, E. 2018.
\newblock Learning explanatory rules from noisy data.
\newblock \emph{Journal of Artificial Intelligence Research}, 61: 1--64.

\bibitem[{Gilmer et~al.(2017)Gilmer, Schoenholz, Riley, Vinyals, and
  Dahl}]{gilmer2017neural}
Gilmer, J.; Schoenholz, S.~S.; Riley, P.~F.; Vinyals, O.; and Dahl, G.~E. 2017.
\newblock Neural message passing for quantum chemistry.
\newblock In \emph{International conference on machine learning}, 1263--1272.
  PMLR.

\bibitem[{Kipf et~al.(2018)Kipf, Fetaya, Wang, Welling, and Zemel}]{Kipf2018}
Kipf, T.; Fetaya, E.; Wang, K.~C.; Welling, M.; and Zemel, R. 2018.
\newblock {Neural Relational Inference for Interacting Systems}.
\newblock In \emph{Proceedings of the 35th International Conference on Machine
  Learning}.

\bibitem[{Kipf and Welling(2016)}]{kipf2016semi}
Kipf, T.~N.; and Welling, M. 2016.
\newblock Semi-supervised classification with graph convolutional networks.
\newblock \emph{arXiv preprint arXiv:1609.02907}.

\bibitem[{Loukas(2019)}]{loukas2019graph}
Loukas, A. 2019.
\newblock What graph neural networks cannot learn: depth vs width.
\newblock \emph{arXiv preprint arXiv:1907.03199}.

\bibitem[{Marcus(2020)}]{marcus2020decade}
Marcus, G. 2020.
\newblock The Next Decade in AI: Four Steps Towards Robust Artificial
  Intelligence.
\newblock arXiv:2002.06177.

\bibitem[{Maron et~al.(2019)Maron, Ben-Hamu, Serviansky, and
  Lipman}]{maron2019provably}
Maron, H.; Ben-Hamu, H.; Serviansky, H.; and Lipman, Y. 2019.
\newblock Provably powerful graph networks.
\newblock \emph{arXiv preprint arXiv:1905.11136}.

\bibitem[{McKay and Piperno(2014{\natexlab{a}})}]{mckay2014practical}
McKay, B.~D.; and Piperno, A. 2014{\natexlab{a}}.
\newblock Practical graph isomorphism, II.
\newblock \emph{Journal of symbolic computation}, 60: 94--112.

\bibitem[{McKay and Piperno(2014{\natexlab{b}})}]{MCKAY201494}
McKay, B.~D.; and Piperno, A. 2014{\natexlab{b}}.
\newblock Practical graph isomorphism, II.
\newblock \emph{Journal of Symbolic Computation}, 60: 94--112.

\bibitem[{Morris et~al.(2019)Morris, Ritzert, Fey, Hamilton, Lenssen, Rattan,
  and Grohe}]{morris2019weisfeiler}
Morris, C.; Ritzert, M.; Fey, M.; Hamilton, W.~L.; Lenssen, J.~E.; Rattan, G.;
  and Grohe, M. 2019.
\newblock Weisfeiler and leman go neural: Higher-order graph neural networks.
\newblock In \emph{Proceedings of the AAAI Conference on Artificial
  Intelligence}, volume~33, 4602--4609.

\bibitem[{Pataki(1998)}]{pataki1998rank}
Pataki, G. 1998.
\newblock On the rank of extreme matrices in semidefinite programs and the
  multiplicity of optimal eigenvalues.
\newblock \emph{Mathematics of operations research}, 23(2): 339--358.

\bibitem[{Qu et~al.(2020)Qu, Chen, Xhonneux, Bengio, and Tang}]{qu2020rnnlogic}
Qu, M.; Chen, J.; Xhonneux, L.-P.; Bengio, Y.; and Tang, J. 2020.
\newblock Rnnlogic: Learning logic rules for reasoning on knowledge graphs.
\newblock \emph{arXiv preprint arXiv:2010.04029}.

\bibitem[{Scarselli et~al.(2009)Scarselli, Gori, Tsoi, Hagenbuchner, and
  Monfardini}]{Scarselli2009}
Scarselli, F.; Gori, M.; Tsoi, A.~C.; Hagenbuchner, M.; and Monfardini, G.
  2009.
\newblock {The Graph Neural Network Model}.
\newblock \emph{IEEE Transactions on Neural Networks}, 20(1): 61--80.

\bibitem[{Shervashidze et~al.(2011)Shervashidze, Schweitzer, {Van Leeuwen},
  Mehlhorn, and Borgwardt}]{Shervashidze2011}
Shervashidze, N.; Schweitzer, P.; {Van Leeuwen}, E.~J.; Mehlhorn, K.; and
  Borgwardt, K.~M. 2011.
\newblock {Weisfeiler-Lehman graph kernels}.
\newblock \emph{Journal of Machine Learning Research}, 12: 2539--2561.

\bibitem[{Veli{\v{c}}kovi{\'c} et~al.(2017)Veli{\v{c}}kovi{\'c}, Cucurull,
  Casanova, Romero, Lio, and Bengio}]{velivckovic2017graph}
Veli{\v{c}}kovi{\'c}, P.; Cucurull, G.; Casanova, A.; Romero, A.; Lio, P.; and
  Bengio, Y. 2017.
\newblock Graph attention networks.
\newblock \emph{arXiv preprint arXiv:1710.10903}.

\bibitem[{Veli{\v{c}}kovi{\'{c}} et~al.(2018)Veli{\v{c}}kovi{\'{c}}, Cucurull,
  Casanova, Romero, Li{\`{o}}, and Bengio}]{Velickovic2018}
Veli{\v{c}}kovi{\'{c}}, P.; Cucurull, G.; Casanova, A.; Romero, A.; Li{\`{o}},
  P.; and Bengio, Y. 2018.
\newblock {Graph Attention Networks}.
\newblock In \emph{Proceeding of the 6th International Conference on Learning
  Representations}, 1--12.

\bibitem[{Wang, Chang, and Kolter(2017)}]{wang2017mixing}
Wang, P.-W.; Chang, W.-C.; and Kolter, J.~Z. 2017.
\newblock The mixing method: coordinate descent for low-rank semidefinite
  programming.

\bibitem[{Wang et~al.(2019)Wang, Donti, Wilder, and Kolter}]{wang2019satnet}
Wang, P.-W.; Donti, P.; Wilder, B.; and Kolter, Z. 2019.
\newblock Satnet: Bridging deep learning and logical reasoning using a
  differentiable satisfiability solver.
\newblock In \emph{International Conference on Machine Learning}, 6545--6554.
  PMLR.

\bibitem[{Wang and Kolter(2019)}]{wang2019low}
Wang, P.-W.; and Kolter, J.~Z. 2019.
\newblock Low-rank semidefinite programming for the MAX2SAT problem.
\newblock In \emph{Proceedings of the AAAI Conference on Artificial
  Intelligence}, volume~33, 1641--1649.

\bibitem[{Weisfeiler and Leman(1968)}]{Weisfeiler1968}
Weisfeiler, B.~Y.; and Leman, A.~A. 1968.
\newblock {The Reduction of a Graph To Canonical Form and the Algebra which
  Appears Therein}.
\newblock 1--11.

\bibitem[{Zhang et~al.(2019)Zhang, Bu, Ester, Zhang, Yao, Yu, and
  Wang}]{zhang2019}
Zhang, Z.; Bu, J.; Ester, M.; Zhang, J.; Yao, C.; Yu, Z.; and Wang, C. 2019.
\newblock {Hierarchical Graph Pooling with Structure Learning}.

\end{thebibliography}


\end{document}